\documentclass{article}
\usepackage{ijcai25}

\usepackage{times}
\usepackage{soul}
\usepackage{url}
\usepackage[hidelinks]{hyperref}
\usepackage[utf8]{inputenc}
\usepackage[small]{caption}
\usepackage{graphicx}
\usepackage{amsmath}
\usepackage{amsthm}
\usepackage{booktabs}
\usepackage{algorithm}
\usepackage{algorithmic}
\usepackage[switch]{lineno}

\usepackage{booktabs}
\usepackage{makecell}
\usepackage{xltabular}
\usepackage{array}
\usepackage{multirow}

\usepackage{fontawesome5}
\usepackage{xcolor}
\definecolor{goldmedal}{RGB}{255, 215, 0}
\definecolor{silvermedal}{RGB}{192, 192, 192}
\definecolor{bronzemedal}{RGB}{205, 127, 50}

\newcommand{\goldmedal}{\textcolor{goldmedal}{\faMedal}}
\newcommand{\silvermedal}{\textcolor{silvermedal}{\faMedal}}
\newcommand{\bronzemedal}{\textcolor{bronzemedal}{\faMedal}}

\title{MME-Industry: A Cross-Industry Multimodal Evaluation Benchmark}

\author{
Dongyi Yi$^1$
\and
Guibo Zhu$^{1,2}$\and
Chenglin Ding$^1$\and
Zongshu Li$^1$\and 
Dong Yi$^{1,2}$ \And Jinqiao Wang$^{1,2}$ \\
\affiliations
$^1$Wuhan AI Research\\
$^2$Institute of Automation, Chinese Academy of Sciences\\
\emails
\{yidongyi, dingchenglin, lizongshu\}@wair.ac.cn,
\{gbzhu, jqwang\}@nlpr.ia.ac.cn,
dong.yi@ia.ac.cn
}

\begin{document}

\maketitle

\begin{abstract}
    With the rapid advancement of Multimodal Large Language Models (MLLMs), numerous evaluation benchmarks have emerged. 
However, comprehensive assessments of their performance across diverse industrial applications remain limited. 
In this paper, we introduce MME-Industry, a novel benchmark designed specifically for evaluating MLLMs in industrial settings.
The benchmark encompasses 21 distinct domain, comprising 1050 question-answer pairs with 50 questions per domain. To ensure data integrity and prevent potential leakage from public datasets, all question-answer pairs were manually crafted and validated by domain experts. 
Besides, the benchmark's complexity is effectively enhanced by incorporating non-OCR questions that can be answered directly, along with tasks requiring specialized domain knowledge. 
Moreover, we provide both Chinese and English versions of the benchmark, enabling comparative analysis of MLLMs' capabilities across these languages. Our findings contribute valuable insights into MLLMs' practical industrial applications and illuminate promising directions for future model optimization research.
\end{abstract}

\section{Introduction}
Recent years we have witnessed rapid advancement in MLLMs\cite{zhang2024mmsurvey}, accompanied by the emergence of multimodal benchmarks for their evaluation. 
These benchmarks serve as crucial instruments for assessing model performance, enabling researchers to systematically analyze MLLMs' capabilities across specific tasks. 
The evolution of evaluation frameworks has paralleled the development of MLLMs themselves, reflecting the growing complexity and sophistication of multimodal understanding and generation capabilities. 
As models continue to advance, the need for robust and comprehensive evaluation methods becomes increasingly critical.
Prior studies\cite{zhang2024mmstar,liu2023mmbench} have established that comprehensive evaluation frameworks are essential for understanding the strengths and limitations of these sophisticated models. 
The rapid iteration of model architectures and training techniques has created a pressing need for standardized evaluation protocols that can effectively measure progress in the field. 
Benchmarks such as MME-Realworld \cite{mme_realworld} is a real-world question-answering (QA) dataset specifically designed for traffic scenarios, DocVQA \cite{mathew2021docvqa} for document understanding, and OCRBench \cite{huang2023ocrbench} for text recognition in images have been developed to address specific tasks within their respective domains. 

However, a comprehensive benchmark tailored for evaluating MLLMs across diverse industry verticals remains notably absent, leaving a critical gap in understanding their applicability in specialized real-world scenarios.
To address this research gap, we introduce a novel benchmark named MME-Industry, specifically designed for industrial applications. The development of this benchmark involved a comprehensive selection process encompassing over 21 distinct industrial sectors, 
including but not limited to the power generation sector, electronics manufacturing, textile production, steel industry, and chemical processing.

To ensure the reliability and accuracy of our benchmark, we engaged domain experts from each respective industry to meticulously annotate and validate the test cases. 
This rigorous validation process by industry professionals guarantees the practical relevance and technical precision of the MME-Industry benchmark.

The dataset comprises a comprehensive collection of 1,050 high-resolution images, each paired with a corresponding question-answer (QA) pair, establishing a robust foundation for evaluation. These images, with an average resolution of 1110 × 859 pixels, exhibit exceptional visual clarity and contain rich, intricate details essential for industrial analysis.
To enhance the challenge level and practical relevance of the evaluation set, we implemented two strategic modifications: \textbf{(1)} the systematic removal of Optical Character Recognition (OCR)-based questions to eliminate text-dependent solutions, and \textbf{(2)} the incorporation of domain-specific questions requiring specialized industrial knowledge. This deliberate design choice ensures that the benchmark effectively assesses the model's capability in handling complex industrial scenarios rather than relying on superficial text recognition.

\begin{figure*}[htbp]
    \centering 
    \includegraphics[width=\textwidth]{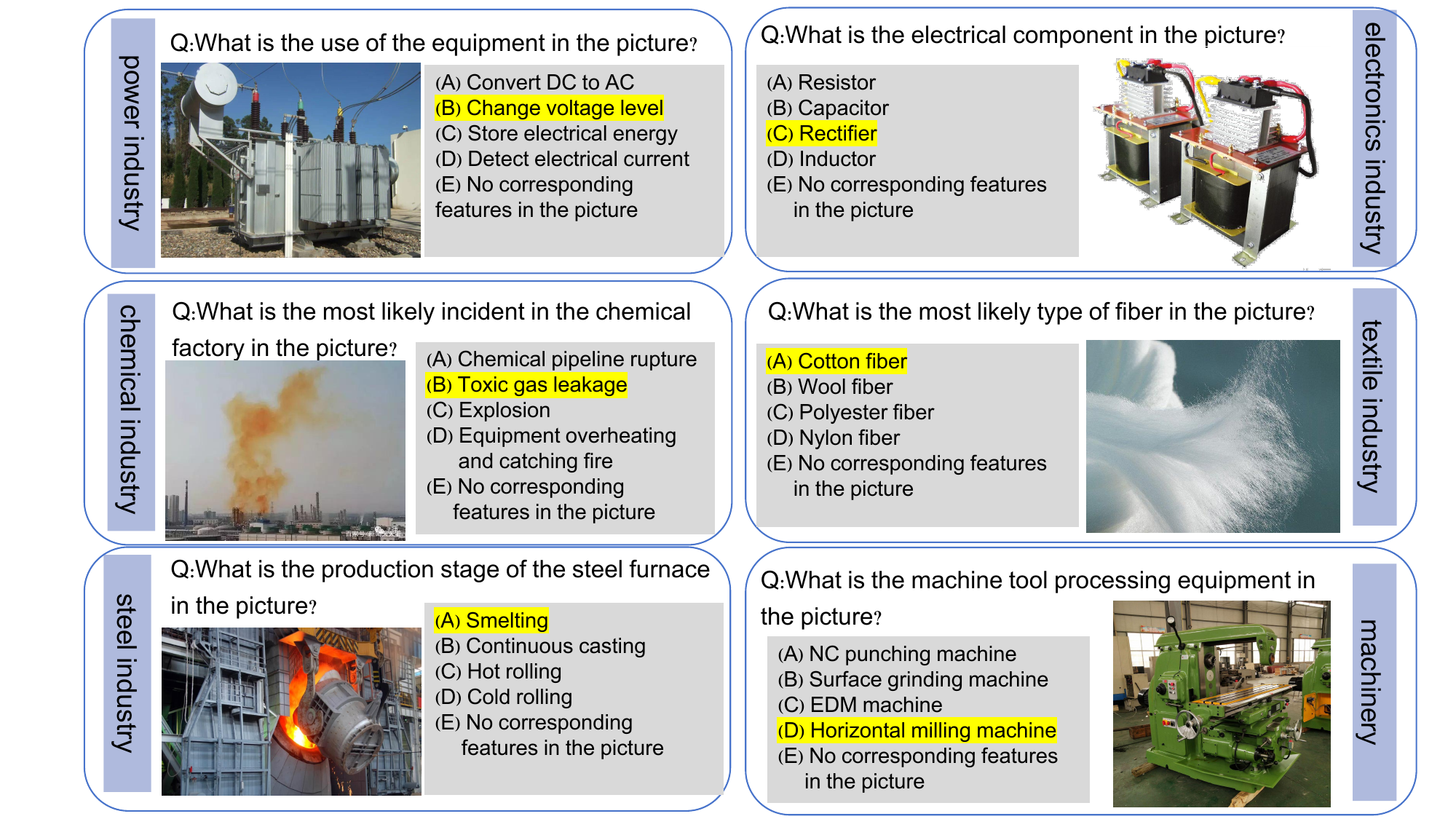} 
    \caption{MME-Industry:
    Our benchmark covers over 20 industries. The diagram shows a subset of the MME Industry Evaluation Set, featuring six sectors: power, electronics, chemical, textile, steel, and machinery. Questions are in a multiple-choice format with five options, highlighting the correct answer, representative of the full evaluation set.} 
    \label{fig:industry_samples} 
\end{figure*}

As illustrated in Figure \ref{fig:industry_samples}, we present six representative industry evaluation scenarios, each demonstrating the benchmark's capacity to address diverse and challenging industrial contexts. These samples showcase the benchmark's comprehensive coverage of various industrial applications and its ability to evaluate model performance across different technical domains.

Our benchmark offers several key advantages:
\begin{itemize}
    \item \textbf{Comprehensive Coverage:} The benchmark covers 21 industrial sectors, encompassing 1,050 QA processes across power, electronics, textiles, steel, chemicals, and more.
    
    \item \textbf{Expert-Validated Content:} Our dataset incorporates 50 test cases per sector, which require domain-specific knowledge and have been verified by industry professionals. The evaluation emphasizes professional reasoning and decision-making capabilities, rather than basic OCR tasks.
    
    \item \textbf{Data Integrity:} Our team of experts manually creates all content, ensuring data integrity and preventing leakage issues that exist in other datasets.
    
    \item \textbf{Multilingual Support:} The benchmark provides both English and Chinese versions (MME-Industry and MME-Industry-CN), supporting cross-lingual research initiatives.
\end{itemize}

The remainder of this paper is organized as follows: Section 2 reviews related work in MLLMs and evaluation benchmarks. 
Section 3 details the construction and characteristics of our MME-Industry benchmark. Section 4 presents our experimental setup 
and results. Section 5 concludes the paper and suggests future research directions.

\section{Related Work}

\subsection{Multimodal Large Language Models}

The emergence of MLLMs is based on the foundation of Large Language Models (LLMs)\cite{chowdhery2023palm,ouyang2022training,team2023internlm,touvron2023llama,yang2023baichuan,bai2023qwen,glm2024chatglm}.
 The pioneering CLIP\cite{radford2021clip} framework demonstrated effective vision-language alignment through contrastive learning\cite{jia2021scaling,radford2021learning}. 
 This breakthrough has inspired various architectural innovations: ranging from the sophisticated gated cross-attention mechanism of Flamingo\cite{awadalla2023openflamingo} to the Q-former design of BLIP-2\cite{li2023blip2}. 
 A parallel trend emerged with MiniGPT-4\cite{zhu2023minigpt4} and LLaVA\cite{liu2024LLaVA}, showcasing the effectiveness of simpler architectural choices using MLPs\cite{taud2018multilayer}. 
 The field has witnessed substantial enhancements in two key areas: multimodal instruction datasets\cite{chen2025sharegpt4v,liu2024improved,wang2023see,ye2023mplug} and novel alignment techniques\cite{bai2023qwenvl,dong2024internvl2,liu2024sphinx,li2024monkey,wang2023cogvlm}. 
 The integration of Low-Rank Adaptation (LoRA)\cite{hu2021lora} techniques has further refined the multimodal understanding capabilities\cite{dong2024internvl2,ye2023mplug}. 
 While commercial APIs such as GPT-4v\cite{achiam2023gpt4} and Gemini-Pro-V\cite{team2023gemini} have made these technologies more accessible, important questions persist regarding their real-world efficacy\cite{mme_realworld,mme}, particularly in complex applications involving models like PaLM-E\cite{driess2023palm} and mPLUG-Owl3\cite{ye2024mplugowl3longimagesequenceunderstanding}.

\subsection{MLLMs Benchmarks}
Recent advancements in MLLMs have significantly propelled the establishment of comprehensive evaluation benchmarks \cite{duan2024vlmevalkit}. These benchmarks systematically assess various dimensions of multimodal understanding: MME \cite{mme} and MME-Realworld \cite{mme_realworld} primarily focus on real-world applications, whereas MMBench \cite{liu2023mmbench} and MMVet \cite{yu2023mm-vet} evaluate detailed visual perception and domain-specific expertise. Furthermore, sophisticated frameworks such as MMStar \cite{zhang2024mmstar} and MMMU \cite{yue2024mmmu} broaden the evaluation spectrum by investigating integrated reasoning abilities across diverse modalities.
Concurrently, specialized evaluation frameworks have been developed to cater to specific domains: ChartQA \cite{masry2022chartqa} is dedicated to chart interpretation, DocVQA \cite{mathew2021docvqa} addresses document analysis, and OCRBench \cite{huang2023ocrbench} measures text recognition accuracy in images. Although these benchmarks have significantly advanced our comprehension of Vision-Language Model (VLM) capabilities, there is a notable deficiency in comprehensive evaluation frameworks that tackle the distinct challenges prevalent across various industrial sectors.
In the domain of Chinese language evaluation, the availability of benchmarks remains notably limited, with only a few datasets such as MME-Realworld-CN \cite{mme_realworld} and MMBench-CN \cite{liu2023mmbench} currently accessible. This highlights a critical area for future research and development to bridge the existing gaps and enhance the robustness of multimodal evaluations in diverse linguistic and industrial contexts.

\section{MME-Industry Benchmark}

\subsection{Dataset Construction}
Our dataset construction process follows a rigorous four-phase approach:

\begin{itemize}
    \item \textbf{Data Collection:} Domain experts curated 50 industry-specific images, with each image paired with a question and four multiple-choice options to ensure content-question alignment.  
    
    \item \textbf{Quality Control:} All images were verified for clarity and relevance, with answers checked for accessibility and compliance with industry standards.  
    
    \item \textbf{Expert Review:} The review process eliminated duplicates and resolved issues, replacing problematic content to maintain industry relevance and professional quality.  
    
    \item \textbf{Translation:} All content, including questions and choices, was accurately translated into English to enhance accessibility for international researchers.  

\end{itemize}

For each image the models need to recoginze, we prepare one question and five options manually, with only one correct answer, three similar wrong options and an option 
``E". This option means the model reject to answer the question because it is unable to recoginze the features of the image or the model's api fail to decode the images.
Additionally, if the image exceeds the input size limit or the model considers the input illegal, the default output is ``E".
Besides, most of our questions require certain professional knowledge or related data storage to work out the questions, which significantly rasies the difficulty of 
our benchmark. All questions are recorded in the annotaions manually. 
The format of the question input(English version, the original version is in Chinese) is shown in Table \ref{question}.

\begin{table}[h]
    \begin{raggedright}
      \caption{Format of the Question}
      \label{question}
        \begin{tabular}{p{0.4\textwidth}}
        \hline
        [Image][Questions] The options: \\
        (A) [Option A] \\
        (B) [Option B] \\
        (C) [Option C] \\
        (D) [Option D] \\
        (E) There are no corresponding features in the image. \\
        Just answer the question based on the letter of the option, 
        your output should only be one letter.\\
        \hline
        \end{tabular}
    \end{raggedright}
\end{table}

\subsection{Task Categories}
To better evaluate the models' value in each application domain, we classify the subtasks by the professional knowledge the questions involve. Here is a brief introduction of the subtasks.

\textbf{Power:}identification and understanding of various electrical power facilities, equipment, and components, focusing on their functions, characteristics, and applications within complex contexts.
\textbf{Electronic:}identification and functional understanding of various electronic circuits, components, and devices, focusing on their roles and characteristics within complex systems. 
\textbf{Textile:}various aspects of textiles, including fiber types, fabric structures, textile machinery, and quality control
\textbf{Steel:}identification and understanding of various stages and processes in steel production, including smelting, rolling, and surface treatment. 
\textbf{Chemical:}identification and understanding of various chemical processes, equipment, and safety measures in the chemical industry.
\textbf{Environmental Protection:}identification and understanding of various environmental processes and technologies, including wastewater treatment, environmental monitoring, ecological restoration, and renewable energy.
\textbf{Machinery:}various industrial manufacturing processes, equipment, and technologies, designed to assess the technical knowledge and reasoning skills of professionals in the manufacturing and engineering fields.
\textbf{Building materials:}identification, analysis, and application of various building materials and construction techniques, 
\textbf{Transportation:}ability to identify and interpret various traffic signs and road conditions, focusing on understanding their meanings and implications for safe driving. 
\textbf{Education:}ability to identify and analyze academic questions and answers across various subjects, focusing on understanding the context and selecting appropriate responses.
\textbf{Financial:}understanding and analysis of various financial concepts, tools, and market trends, including stock identification, financial statements, and monetary policy.
\textbf{Agricultural:}comprehensively evaluates various aspects of agricultural practices, technologies, and management, including irrigation systems, agricultural machinery, crop management, and sustainable farming techniques.
\textbf{Automobile:}evaluates various aspects of automotive technology, components, and services, including electric vehicle batteries, autonomous driving sensors, manufacturing processes, and safety systems.
\textbf{Light:}various aspects of manufacturing processes and technologies across different industries, including ceramics, textiles, plastics, paper, and food processing.
\textbf{Petrochemical:}evaluation of the petrochemical sector, covering key areas such as oil extraction, safety management, maintenance, pipeline inspection, emergency response, chemical production, and automation, through visual recognition and classification tasks.
\textbf{Food:}various food production and processing sectors, including seafood, dairy, grain, wine, tea, coffee, canned food, baking, oil production, and specialty food items, through visual recognition and classification tasks.
\textbf{Cultural and tourism:}identification and reasoning skills related to various global landmarks and tourist attractions, covering their recognition, geographical locations, historical backgrounds, and cultural significance.
\textbf{Entertainment:}character identification and reasoning, covering various notable figures from different fields such as politics, entertainment, science, and literature. ability to recognize individuals and understand their nationalities, professions, works, and contributions.
\textbf{Medical:}medical knowledge and reasoning, covering various aspects such as identifying human body parts, diagnosing diseases, understanding organ functions, and evaluating the effects and side effects of medications. It tests the ability to recognize and analyze medical images and information.
\textbf{Gaming:}various aspects of gaming technology, including game design, development, and distribution, focusing on understanding the technical aspects of game development and design.
\textbf{Nonferrous:}knowledge and reasoning skills related to minerals and metals, focusing on identifying their main components, properties, uses, and applications in various industries. It tests the ability to recognize and analyze characteristics of different materials.

\subsection{Evaluation Metrics}
Although we ask the models to answer with only one letter for each question, the MLLMs may still answer with extra information such as explanatory texts or brackets. 
We consider the ability to follow instructions as an essential aspect of MLLMs evaluation. Therefore, we directly use the model's complete response as the final answer without additional extraction. Performance scores are reported as percentages, calculated using the following formula:

$$ S = \frac{\sum_{i=1}^{N} s_i}{N} \times 100\% $$
where $S$ denotes the overall accuracy (in percentage), $s_i \in \{0,1\}$ denotes the individual score for the i-th question, and $N$ is the total number of questions in the benchmark.

\section{Experiments}

\subsection{Experimental Setup}
We performed extensive evaluations of multiple state-of-the-art multimodal large language models (MLLMs) on the proposed MME-Industry benchmark. The experimental configuration is detailed below:

\subsubsection{Model Selection}
We evaluated state-of-the-art models from the OpenCompass multimodal ranking\footnote{\url{https://rank.opencompass.org.cn/leaderboard-multimodal/?m=24-12}} (as of December 30, 2024), focusing on the top 10 performers. From the InternVL2.5 family, which includes four variants (InternVL2.5-78B-MPO, InternVL2.5-38B-MPO, InternVL2.5-78B, and InternVL2.5-38B), we selected InternVL2.5-78B-MPO due to its superior performance.
Our evaluation also encompassed other leading models: GPT-4o, Claude-3.5-sonnet, Gemini-1.5-Flash-exp-0827, MiniCPM-V-2.6, and GLM-4V-Flash. In total, we assessed 10 models\footnote{Taiyi and TelemM were excluded due to API unavailability.}. The evaluated models include: SenseChat-Vision, InternVL2.5-78B, Qwen2-VL-72B-Instruct, JT-VL-Chat-V3.0, Step-1.5V, GPT-4o, Claude-3.5-Sonnet, Gemini-1.5-Flash-exp-0827, GLM-4V-Flash, and MiniCPM-V-2.6.

\subsubsection{Parameter Settings}
For consistent evaluation across all models, we standardized the following parameters: temperature (0.2), maximum output length (64 tokens), and top-p sampling (0.95).

Due to image size constraints, the Qwen2-VL-72B-Instruct model\footnote{\url{https://huggingface.co/Qwen/Qwen2-VL-72B-Instruct}} processes a default range of 4 to 16,384 visual tokens per image. In our experiments, we configured the input resolution with a minimum of 
$256 \times 28 \times 28$ pixels and a maximum of $1280 \times 28 \times 28$ pixels.

\subsection{Results and Analysis}

\begin{figure*}[ht]
    \centering
    \includegraphics[width=0.5\textwidth]{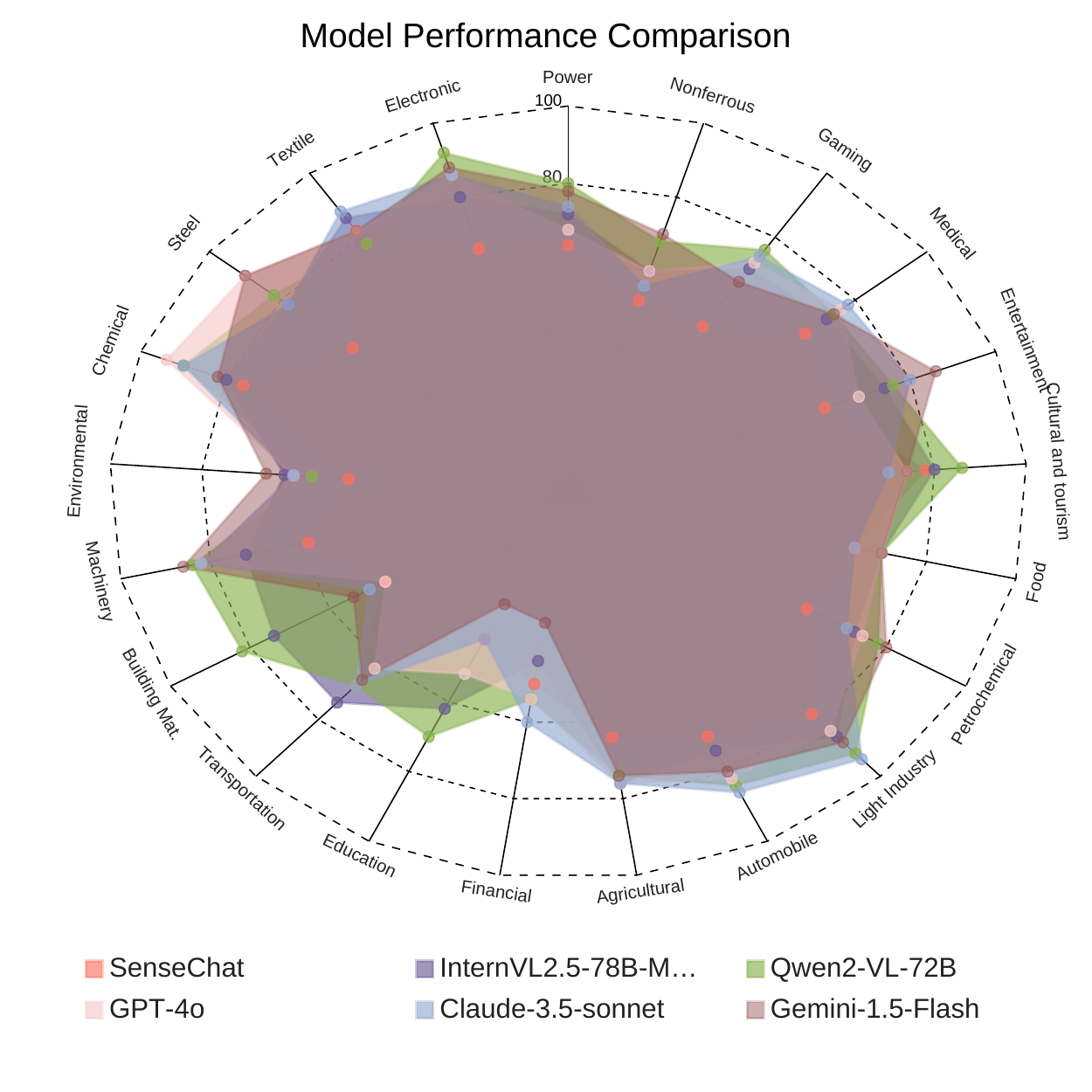} 
    \caption{Comparison of 6 top MLLMs across 21 industries in English evaluate. The maximum score for each industry is 100.} 
    \label{fig:model_performance}
\end{figure*}

\subsubsection{Results}
    Based on the comparative analysis shown in Table \ref{tab:model_scores}, Qwen2-VL-72B-Instruct demonstrates superior performance across both Chinese and English evaluations, achieving the highest scores of 78.66\% and 75.04\% respectively. 
Claude-3.5-Sonnet consistently ranks second, with scores of 74.09\% for Chinese and 72.66\% for English tasks. InternVL2.5-78B-MPO shows strong performance in Chinese evaluation (73.42\%), 
while Gemini-1.5-Flash maintains competitive performance in both languages (72.76\% CN, 72.38\% EN). Notably, while most models maintain relatively consistent performance across languages, 
some models show significant variations between Chinese and English capabilities. MiniCPM-V-2.6, for instance, shows the largest performance gap, scoring substantially lower in Chinese (18.47\%) compared to English (29.04\%). The results indicate that large-scale models generally demonstrate more robust and balanced cross-lingual capabilities.
\begin{table}[ht]
    \centering
    \begin{tabular}{l r r r}
        \toprule
        \textbf{Model} & \textbf{CN-Score}  & \textbf{EN-Score} & \textbf{Size}\\
        \midrule
        Qwen2-VL-72B-Instruct & \goldmedal 78.66\%  & \goldmedal 75.04\% & 72B \\
        Claude-3.5-Sonnet & \silvermedal 74.09\%  & \silvermedal 72.66\% & -\\
        InternVL2.5-78B-MPO & \bronzemedal 73.42\%  & 70.95\% & 78B \\
        Gemini-1.5-Flash & 72.76\% & \bronzemedal 72.38\% & - \\
        SenseChat-Vision & 70.38\%  & 62.09\% & - \\
        GPT-4o & 70.28\%  & 72.00\% & - \\
        GLM-4V-Flash & 65.61\%  & 64.38\% & - \\
        JT-VL-Chat-V3.0 & 58.85\%  & 66.09\% & - \\
        Step-1.5V & 52.00\%  & 63.23\% & - \\
        MiniCPM-V-2.6 & 18.47\%  & 29.04\% & 8B \\
        \bottomrule
    \end{tabular}
    \caption{Comparison of Model Scores with image input.}
    \label{tab:model_scores}
\end{table}

To ensure the absence of data leakage, we conducted a test by evaluating the model's scores in the absence of image inputs. The results are presented in Table \ref{tab:model_scores_no_image}, revealing an average CN-Score of 16.46\% and an average EN-Score of 16.78\%. 
These values are notably lower than the threshold of 25\%, which corresponds to the performance level expected from randomly selected answers. This finding confirms that the model is not relying on data leakage to achieve its performance.
\begin{table}
    \centering
    \begin{tabular}{l r r r}
        \toprule
        \textbf{Model} & \textbf{CN-Score}  & \textbf{EN-Score} & \textbf{Size} \\
        \midrule
        SenseChat-Vision &  17.33\%  &  12.38\% & - \\  
        InternVL2.5-78B-MPO & \bronzemedal 23.90\%  &  20.95\% & 78B \\
        Qwen2-VL-72B-Instruct &  16.38\%  & 15.62\% & 72B \\ 
        JT-VL-Chat-V3.0 & \silvermedal 30.76\% &  \silvermedal 32.76\% & - \\
        Step-1.5V & 18.57\%  & \goldmedal 33.81\% & - \\
        GPT-4o & 6.47\%  & 6.67\% & - \\
        Claude-3.5-Sonnet & 17.23\%  & 17.14\% & - \\
        Gemini-1.5-Flash & 0.00\%  & 0.00\% & - \\
        GLM-4V-Flash & \goldmedal 32.28\%  & \bronzemedal 29.04\% & - \\
        MiniCPM-V-2.6 & 0.00\%  & 14.85\% & 8B \\
        \midrule
        Average score & 16.46\% & 16.78\% \\
        \bottomrule
    \end{tabular}
    \caption{Comparison of Model Scores without image input.}
    \label{tab:model_scores_no_image}
\end{table}

Table \ref{tab:performance-comparison-CN} shows the performance of 21 industries in Chinese evaluation.
In the technical fields, Qwen2 VL-72B demonstrates exceptional performance, achieving accuracy rates of 94\% in the Electronics domain, 92\% in the Light Industry domain, and 90\% in the Chemical domain. 
Claude 3.5-Sonnet maintains consistently high performance across most technical industries, with scores exceeding 80\% in nearly all areas. 
GPT-4o also exhibits strong capabilities in technical sectors, particularly in the Electronics domain (88\%) and the Light Industry domain (86\%).
Among the high-performance domains are Electronics, where most models achieve accuracy rates above 80\%. Light Industry, with an average performance exceeding 80\%. and the Chemical domain, which reaches a peak performance of 94\%. 
In contrast, the Financial domain, Education domain, and Building Materials domain are identified as challenging areas, with average scores below 50\%. The Environmental sector exhibits significant variability, with model performance ranging from 38\% to 66\%.
JT-VL Chat-V3 shows a substantial reliance on visual input, as evidenced by the large gaps between its image-inclusive and image-exclusive scores. 
Meanwhile, MiniCPM V-2.6 demonstrates limited overall performance, suggesting challenges in understanding the Chinese industrial domain.

\begin{table*}[t]
    \centering
    \small
    \setlength{\tabcolsep}{4pt}
    \begin{tabular}{l*{10}{c}}
    \toprule
    \textbf{Industry} & 
    \makecell{SenseChat\\Vision} & 
    \makecell{InternVL2.5\\78B-MPO} & 
    \makecell{Qwen2\\VL-72B} & 
    \makecell{JT-VL\\Chat-V3} & 
    \makecell{Step\\1.5V} & 
    \makecell{GPT-4o} & 
    \makecell{Claude\\3.5-Sonnet} &
    \makecell{Gemini\\1.5-Flash} & 
    \makecell{GLM-4v\\Flash} & 
    \makecell{MiniCPM\\V-2.6} \\
    \midrule
    \multicolumn{11}{l}{\textcolor{red}{*}\textbf{ The score shown in parentheses excludes image input.}} \\
    \midrule
    Power          & 78\%(16\%) & 84\%(22\%) & \textcolor{red}{86\%}(30\%) & 62\%(36\%) & 52\%(16\%) & 82\%(4\%) & 82\%(22\%) & 80\%(0\%) & 76\%(44\%) & 28\%(2\%) \\
    Electronic     & 80\%(14\%) & 88\%(18\%) & \textcolor{red}{94\%}(24\%) & 72\%(32\%) & 60\%(12\%) & 88\%(4\%) & 92\%(10\%) & 90\%(0\%) & 74\%(34\%) & 16\%(2\%) \\
    Textile        & 82\%(6\%)  & 80\%(18\%) & 84\%(6\%) & 60\%(16\%) & 54\%(4\%) & 82\%(0\%) & \textcolor{red}{86\%}(12\%) & 84\%(0\%) & 84\%(24\%) & 28\%(0\%) \\
    Steel          & 80\%(16\%) & 72\%(26\%) & 82\%(10\%) & 56\%(32\%) & 60\%(8\%) & \textcolor{red}{84\%}(2\%) & 82\%(26\%) & 82\%(0\%) & 78\%(30\%) & 24\%(0\%) \\
    Chemical       & 78\%(20\%) & 88\%(38\%) & 90\%(16\%) & 76\%(36\%) & 68\%(26\%) & 90\%(2\%) & \textcolor{red}{94\%}(12\%) & 82\%(0\%) & 78\%(40\%) & 30\%(0\%) \\
    \addlinespace
    Environmental  & \textcolor{red}{66\%}(30\%) & 58\%(28\%) & 64\%(26\%) & 48\%(36\%) & 38\%(20\%) & 48\%(16\%) & 64\%(32\%) & 62\%(0\%) & 50\%(38\%) & 0\%(0\%) \\
    Machinery      & 74\%(8\%) & 76\%(16\%) & \textcolor{red}{92\%}(8\%) & 60\%(20\%) & 68\%(12\%) & 78\%(0\%) & 84\%(16\%) & 86\%(0\%) & 88\%(36\%) & 34\%(4\%) \\
    Building Mat.  & 62\%(24\%) & 60\%(22\%) & \textcolor{red}{64}\%(8\%) & 50\%(26\%) & 48\%(30\%) & 48\%(0\%) & 52\%(14\%) & 54\%(0\%) & 54\%(42\%) & 14\%(0\%) \\
    Transportation & 72\%(8\%) & 72\%(26\%) & 70\%(16\%) & 56\%(34\%) & 26\%(18\%) & \textcolor{red}{74\%}(2\%) & 60\%(8\%) & \textcolor{red}{74\%}(0\%) & 58\%(32\%) & 20\%(4\%) \\
    Education      & 48\%(0\%) & 62\%(0\%) & \textcolor{red}{70\%}(8\%) & 48\%(8\%) & 24\%(8\%) & 32\%(0\%) & 50\%(2\%) & 36\%(0\%) & 38\%(10\%) & 8\%(0\%) \\
    \addlinespace
    Financial      & 44\%(18\%) & 48\%(24\%) & \textcolor{red}{62\%}(10\%) & 32\%(42\%) & 48\%(12\%) & 50\%(4\%) & 56\%(34\%) & 26\%(0\%) & 56\%(38\%) & 0\%(0\%) \\
    Agricultural   & 78\%(18\%) & 74\%(38\%) & \textcolor{red}{84\%}(22\%) & 70\%(32\%) & 50\%(16\%) & 78\%(0\%) & 78\%(4\%) & 82\%(0\%) & 78\%(30\%) & 34\%(0\%) \\
    Automobile     & 78\%(10\%) & 74\%(26\%) & 82\%(10\%) & 46\%(18\%) & 48\%(26\%) & 82\%(4\%) & \textcolor{red}{84}\%(2\%) & 80\%(0\%) & 70\%(36\%) & 36\%(16\%) \\
    Light Industry & 84\%(10\%) & 86\%(26\%) & \textcolor{red}{92\%}(2\%) & 56\%(22\%) & 52\%(18\%) & 86\%(0\%) & \textcolor{red}{92\%}(8\%) & 88\%(0\%) & 88\%(18\%) & 48\%(0\%) \\
    Petrochemical  & 52\%(18\%) & \textcolor{red}{78\%}(28\%) & 74\%(12\%) & 58\%(38\%) & 52\%(20\%) & 68\%(0\%) & 74\%(16\%) & 72\%(0\%) & 48\%(32\%) & 12\%(2\%) \\
    \addlinespace
    Food           & 70\%(28\%) & 72\%(22\%) & 70\%(16\%) & 60\%(38\%) & 62\%(24\%) & 68\%(16\%) & 72\%(20\%) & \textcolor{red}{78\%}(0\%) & 60\%(40\%) & 18\%(2\%) \\
    Cultural       & 78\%(10\%) & 80\%(20\%) & \textcolor{red}{90\%}(28\%) & 74\%(26\%) & 78\%(22\%) & 82\%(6\%) & 76\%(22\%) & 82\%(0\%) & 80\%(46\%) & 22\%(0\%) \\
    Entertainment  & 76\%(16\%) & 82\%(12\%) & \textcolor{red}{88\%}(14\%) & 74\%(30\%) & 56\%(8\%) & 72\%(12\%) & 82\%(14\%) & 86\%(0\%) & 66\%(22\%) & 6\%(0\%) \\
    Medical        & 72\%(38\%) & 76\%(36\%) & \textcolor{red}{82\%}(34\%) & 76\%(58\%) & 58\%(30\%) & 72\%(32\%) & 72\%(36\%) & 72\%(0\%) & 48\%(38\%) & 8\%(0\%) \\
    Gaming         & 66\%(20\%) & \textcolor{red}{74\%}(20\%) & \textcolor{red}{74\%}(20\%) & 54\%(18\%) & 48\%(20\%) & 64\%(10\%) & 62\%(16\%) & 64\%(0\%) & 60\%(30\%) & 6\%(0\%) \\
    Nonferrous     & 60\%(36\%) & 58\%(36\%) & 58\%(24\%) & 60\%(48\%) & 54\%(40\%) & 60\%(22\%) & 62\%(36\%) & \textcolor{red}{68\%}(0\%) & 46\%(28\%) & 2\%(0\%) \\
    \bottomrule
    \end{tabular}
    \caption{\textbf{CN-Score:} The evaluation results across 21 industries in the Chinese evaluate indicate that the score highlighted in red represents the highest score achieved by the model in each respective field.}
    \label{tab:performance-comparison-CN}
\end{table*}

Table \ref{tab:performance-comparison-EN} shows the performance of 21 industries in English evaluation.
Qwen2 VL-72B maintains strong performance overall, though it exhibits slightly lower scores compared to CN-Score. In contrast, Claude 3.5-Sonnet demonstrates more consistent performance across various domains in English. Notably, models across the board show robust performance in the Light Industry and Electronic sectors.
The Environmental sector generally exhibits lower performance compared to CN-Score. However, there is more consistent performance across models in technical domains. Additionally, the gaps between image-inclusive and image-exclusive scores are smaller than those observed in CN-Score.
Technical domains maintain high performance, with the Electronic sector peaking at 92\% and the Light Industry sector reaching 94\%. In contrast, the Education and Financial sectors consistently show lower performance across all models. The Medical domain exhibits higher variance across models compared to CN-Score.

\begin{table*}[h]
    \centering
    \small
    \setlength{\tabcolsep}{4pt}
    \begin{tabular}{l*{10}{c}}
    \toprule
    \textbf{Industry} & 
    \makecell{SenseChat\\Vision} & 
    \makecell{InternVL2.5\\78B-MPO} & 
    \makecell{Qwen2\\VL-72B} & 
    \makecell{JT-VL\\Chat-V3} & 
    \makecell{Step\\1.5V} & 
    \makecell{GPT-4o} & 
    \makecell{Claude\\3.5-Sonnet} &
    \makecell{Gemini\\1.5-Flash} & 
    \makecell{GLM-4v\\Flash} & 
    \makecell{MiniCPM\\V-2.6} \\
    \midrule
    \multicolumn{11}{l}{\textcolor{red}{*}\textbf{ The score shown in parentheses excludes image input.}} \\
    \midrule
    Power          & 64\%(2\%) & 72\%(12\%) & \textcolor{red}{80\%}(14\%) & 66\%(26\%) & 66\%(34\%) & 68\%(2\%) & 74\%(0\%) & 78\%(0\%) & 74\%(38\%) & 32\%(10\%) \\
    Electronic     & 66\%(8\%) & 80\%(14\%) & \textcolor{red}{92\%}(10\%) & 70\%(36\%) & 76\%(28\%) & 86\%(6\%) & 86\%(0\%) & 88\%(0\%) & 74\%(32\%) & 36\%(10\%) \\
    Textile        & 82\%(6\%) & 86\%(12\%) & 78\%(8\%) & 78\%(32\%) & 80\%(38\%) & 82\%(0\%) & \textcolor{red}{88\%}(0\%) & 82\%(0\%) & 82\%(13\%) & 21\%(4\%) \\
    Steel          & 60\%(6\%) & 78\%(16\%) & 82\%(8\%) & 68\%(26\%) & 62\%(30\%) & \textcolor{red}{90\%}(0\%) & 78\%(2\%) & \textcolor{red}{90\%}(0\%) & 80\%(16\%) & 44\%(18\%) \\
    Chemical       & 76\%(14\%) & 80\%(34\%) & 90\%(22\%) & 80\%(42\%) & 74\%(50\%) & \textcolor{red}{94\%}(2\%) & 90\%(0\%) & 82\%(0\%) & 78\%(44\%) & 20\%(12\%) \\
    \addlinespace
    Environmental  & 48\%(26\%) & 62\%(32\%) & 56\%(26\%) & 62\%(38\%) & 50\%(34\%) & 60\%(10\%) & 60\%(6\%) & \textcolor{red}{66\%}(0\%) & 38\%(32\%) & 22\%(18\%) \\
    Machinery      & 58\%(2\%) & 72\%(20\%) & 84\%(10\%) & 68\%(30\%) & 82\%(30\%) & 82\%(0\%) & 82\%(0\%) & \textcolor{red}{86\%}(0\%) & \textcolor{red}{86\%}(24\%) & 26\%(14\%) \\
    Building Mat.  & 46\%(2\%) & 46\%(12\%) & 44\%(6\%) & 50\%(26\%) & \textcolor{red}{58\%}(38\%) & 46\%(4\%) & 50\%(0\%) & 54\%(0\%) & 54\%(36\%) & 28\%(12\%) \\
    Transportation & 68\%(4\%) & \textcolor{red}{74\%}(14\%) & 68\%(6\%) & 58\%(26\%) & 40\%(30\%) & 62\%(2\%) & 68\%(6\%) & 66\%(0\%) & 58\%(32\%) & 24\%(20\%) \\
    Education      & 48\%(0\%) & 62\%(0\%) & \textcolor{red}{70\%}(0\%) & 40\%(2\%) & 36\%(8\%) & 52\%(2\%) & 48\%(0\%) & 32\%(0\%) & 36\%(8\%) & 22\%(26\%) \\
    \addlinespace
    Financial      & 50\%(34\%) & 44\%(32\%) & 54\%(22\%) & 44\%(40\%) & 50\%(38\%) & 54\%(14\%) & \textcolor{red}{60\%}(4\%) & 34\%(0\%) & 46\%(38\%) & 24\%(22\%) \\
    Agricultural   & 64\%(8\%) & 76\%(22\%) & 64\%(14\%) & \textcolor{red}{82\%}(26\%) & 68\%(26\%) & 76\%(2\%) & 76\%(0\%) & 74\%(0\%) & 72\%(32\%) & 26\%(14\%) \\
    Automobile     & 70\%(8\%) & 74\%(20\%) & 84\%(24\%) & 72\%(40\%) & 68\%(44) & 82\%(6\%) & \textcolor{red}{86\%}(0\%) & 80\%(0\%) & 74\%(38\%) & 48\%(28\%) \\
    Light Industry & 78\%(10\%) & 86\%(16\%) & 92\%(10\%) & 84\%(24\%) & 78\%(32\%) & 84\%(0\%) & \textcolor{red}{94\%}(4\%) & 88\%(0\%) & 82\%(24\%) & 48\%(12\%) \\
    Petrochemical  & 60\%(12\%) & 72\%(24\%) & 78\%(26\%) & 56\%(38\%) & 64\%(38\%) & 74\%(4\%) & 70\%(0\%) & \textcolor{red}{80\%}(0\%) & 58\%(36\%) & 22\%(14\%) \\
    \addlinespace
    Food           & 64\%(14\%) & \textcolor{red}{70\%}(24\%) & \textcolor{red}{70\%}(18\%) & 56\%(34\%) & 62\%(42\%) & \textcolor{red}{70\%}(12\%) & 64\%(0\%) & \textcolor{red}{70\%}(0\%) & 54\%(30\%) & 32\%(16\%) \\
    Cultural       & 78\%(4\%) & \textcolor{red}{86\%}(18\%) & 74\%(18\%) & 74\%(38\%) & 72\%(50\%) & 74\%(6\%) & 74\%(0\%) & 82\%(0\%) & 76\%(34\%) & 28\%(22\%) \\
    Entertainment  & 60\%(10\%) & 74\%(8\%) & 76\%(6\%) & 80\%(30\%) & 68\%(22\%) & 68\%(6\%) & 80\%(0\%) & \textcolor{red}{86\%}(0\%) & 76\%(8\%) & 10\%(6\%) \\
    Medical        & 66\%(36\%) & 72\%(38\%) & 74\%(24\%) & 72\%(54\%) & 60\%(32\%) & 76\%(30\%) & 78\%(0\%) & 74\%(0\%) & \textcolor{red}{92\%}(22\%) & 20\%(8\%) \\
    Gaming         & 52\%(30\%) & 70\%(34\%) & \textcolor{red}{76\%}(26\%) & 68\%(34\%) & 60\%(30\%) & 72\%(16\%) & 74\%(14\%) & 66\%(0\%) & 62\%(36\%) & 34\%(8\%) \\
    Nonferrous     & 52\%(24\%) & 60\%(38\%) & 68\%(30\%) & 60\%(46\%) & 54\%(36\%) & 60\%(16\%) & 56\%(0\%) & \textcolor{red}{70\%}(0\%) & 46\%(24\%) & 32\%(18\%) \\
    \bottomrule
    \end{tabular}
    \caption{\textbf{EN-Score:} The evaluation results across 21 industries in the English evaluate indicate that the score highlighted in red represents the highest score achieved by the model in each respective field.}
    \label{tab:performance-comparison-EN}
\end{table*}

\subsubsection{Analysis}

The experimental results reveal significant variations in model performance across industrial domains and languages, with several key insights emerging:

\textbf{Cross-Language Performance:} The Qwen2-VL-72B model demonstrates consistent superiority in both Chinese and English tasks, particularly excelling in technical domains. 
For example, it achieves 94\% accuracy in the Electronic sector (Chinese) and 92\% (English), highlighting its robust bilingual capabilities. Similarly, Claude-3.5-Sonnet and GPT-4o exhibit stable cross-lingual performance, with Claude-3.5-Sonnet achieving 94\% accuracy in both the Chemical domain (Chinese) and Light Industry (English), underscoring its effectiveness in multilingual understanding.

\textbf{Domain-Specific Performance:} Models generally perform better in technical and industrial sectors (e.g., Electronic, Chemical, Light Industry) compared to abstract or creative domains (e.g., Cultural, Entertainment). In the Electronic domain, most models achieve above 80\% accuracy, with Qwen2-VL-72B leading at 94\% (Chinese) and 92\% (English). 
Conversely, the Educational and Financial sectors show consistently lower performance across all models, while the Environmental domain presents significant challenges, with performance drops observed in both languages.

\textbf{Model-Specific Insights:} Qwen2-VL-72B stands out for its exceptional capabilities, maintaining high performance across diverse domains with minimal degradation between Chinese and English tasks. 
Its strength in technical and industrial applications is particularly notable. In contrast, MiniCPM-V-2.6, while showing relatively lower performance compared to larger models, demonstrates potential for specific industrial applications and consistent cross-lingual performance, suggesting its utility in resource-constrained scenarios. 

\textbf{Multi-modal Integration:} The scores in parentheses (excluding image input) reveal that visual information significantly enhances model performance, particularly in complex technical domains where visual context aids understanding. 
For instance, SenseChat-Vision's performance in the Power sector drops from 78\% to 16\% without image input, emphasizing the critical role of multi-modal learning in industrial applications.

\textbf{Research Implications:} The performance disparity across domains suggests the need for domain-specific fine-tuning strategies.
Further investigation is needed to understand the models' lower performance in educational and financial domains.
The role of visual information processing capabilities warrants deeper exploration, particularly in technical domains where visual context significantly improves performance.

\section{Conclusion}
In this paper, we introduce MME-Industry, a novel and comprehensive benchmark specifically designed to evaluate the performance of Multimodal Large Language Models (MLLMs) in industrial applications. 
MME-Industry features a diverse set of 1,050 evaluation questions across 21 different industries, covering a wide range of practical scenarios and requirements. This benchmark aims to provide a thorough assessment of the capabilities and limitations of MLLMs when applied to real-world industrial tasks.
We have conducted a thorough evaluation of 10 advanced MLLMs using MME-Industry. The results provide a comprehensive understanding of MLLMs' current capabilities in industrial applications. Our future work will focus on four key directions: (1) expanding the dataset scale to enhance coverage and diversity, (2) increasing the number of tested models to ensure comprehensive evaluation, (3) establishing open-source platforms to foster community engagement, and (4) implementing continuous evaluation and iterative update mechanisms to maintain alignment with the rapid evolution of industrial AI technologies.


\bibliographystyle{named}
\bibliography{ijcai25}

\end{document}